\documentclass[11pt]{article}
\usepackage{coling2020}
\usepackage{times}
\usepackage{url}
\usepackage{latexsym}
\usepackage{bm}
\usepackage{subfigure}
\usepackage{xspace}

\newcommand{\mname}{\texttt{GraSSLM}\xspace}
\usepackage{graphicx}
\usepackage{multirow}
\usepackage{tabularx}

\usepackage{amsmath}
\usepackage{helvet}
\usepackage{courier}
\usepackage{booktabs}
\usepackage{latexsym}
\usepackage{algorithm}
\usepackage{algorithmic}
\usepackage{graphicx}
\usepackage{tabularx}
\usepackage{color}

\usepackage{enumerate}
\usepackage{enumitem}
\usepackage{color}
\usepackage{booktabs}

\newlist{mylist}{enumerate*}{1}
\setlist[mylist]{label=(\roman*)}
\usepackage{lipsum} 
\usepackage{tikz}

\usepackage{amsmath,amssymb,bbm,mathtools,bbold}
\usepackage[symbol]{footmisc}

\colingfinalcopy

\title{A Graph Representation of Semi-structured Data for Web Question Answering }

\author{
  {Xingyao Zhang}$^1$\footnotemark[2] \quad {Linjun Shou}$^2$\footnotemark[2] \quad {Jian Pei}$^3$\footnotemark[3] \quad {Ming Gong}$^2$ \quad \textbf{Lijie Wen}$^1$ \quad \textbf{Daxin Jiang}$^2$\footnotemark[4]\\
  $^1${Tsinghua University} \\
  $^2${STCA NLP Group, Microsoft} \\
  $^3${School of Computing Science, Simon Fraser University} \\
  \small{xingyaozhangthu@gmail.com},
  \small{\{lisho,migon,djiang\}@microsoft.com},\\ \small{jpei@cs.sfu.ca},
  \small{wenlj@tsinghua.edu.cn} \\
}

\date{}

\begin{document}

\maketitle

\footnotetext[2]{Equal contribution. Work done when the first author was an intern at Microsoft STCA.}
\footnotetext[3]{Jian Pei's research is supported in part by the NSERC Discovery Grant program. All opinions, findings, conclusions and recommendations in this paper are those of the authors and do not necessarily reflect the views of the funding agencies.}
\footnotetext[4]{Corresponding author.}

\begin{abstract}
The abundant semi-structured data on the Web, such as HTML-based tables and lists, provide commercial search engines a rich information source for question answering (QA). Different from plain text passages in Web documents, Web tables and lists have inherent structures, which carry semantic correlations among various elements in tables and lists. Many existing studies treat tables and lists as flat documents with pieces of text and do not make good use of semantic information hidden in structures. In this paper, we propose a novel graph representation of Web tables and lists based on a systematic categorization of the components in semi-structured data as well as their relations. We also develop pre-training and reasoning techniques on the graph model for the QA task. Extensive experiments on several real datasets collected from a commercial engine verify the effectiveness of our approach. Our method improves F1 score by 3.90 points over the state-of-the-art baselines.
\end{abstract}

\section{Introduction}
\label{intro}

\blfootnote{
    %
    %
    %
    %
    %
    \hspace{-0.65cm}  
    This work is licensed under a Creative Commons 
    Attribution 4.0 International License.
    License details:
    \url{http://creativecommons.org/licenses/by/4.0/}.
}

\begin{figure}[h!]
\centering
\includegraphics[scale=0.25]{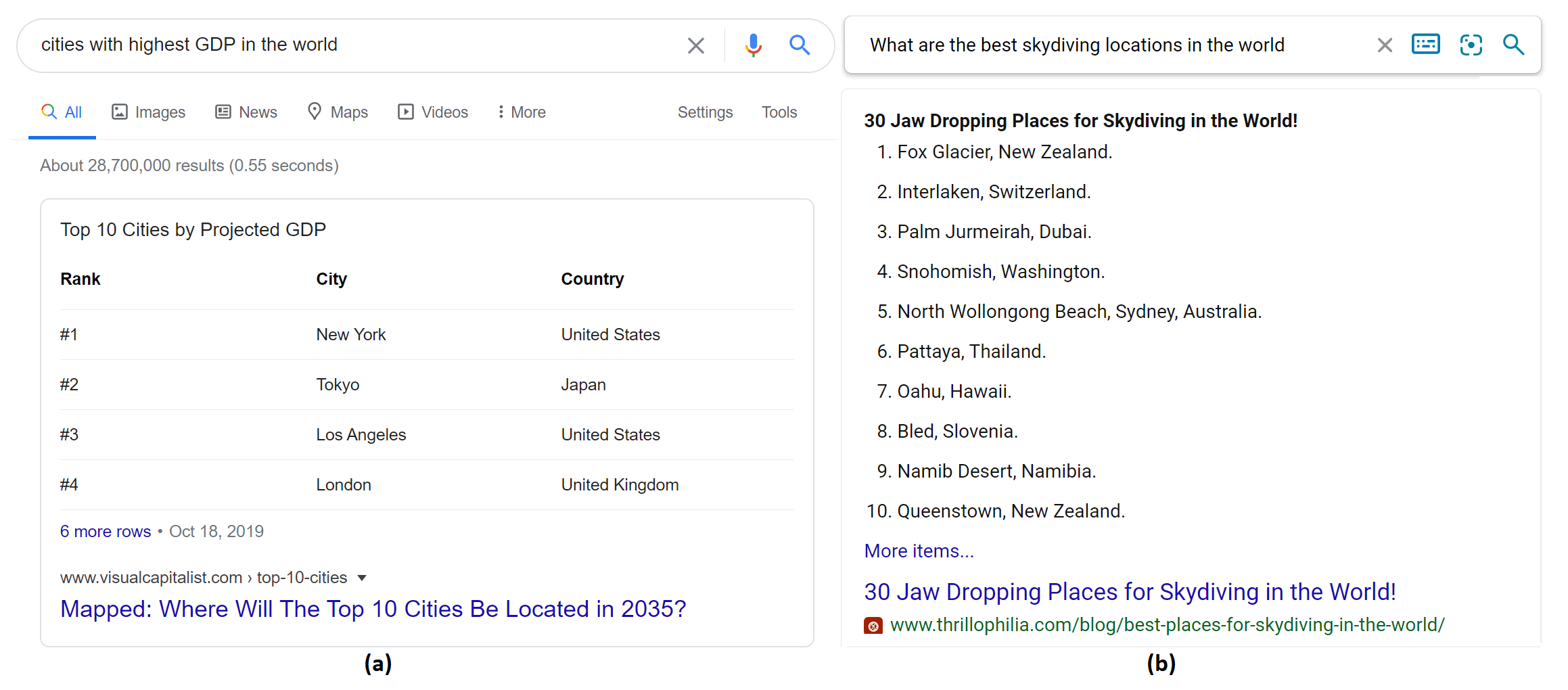}
\vskip -1em
\caption{Examples of Web table (a) and Web list (b) from a commercial search engine. Two queries are non-factoid queries, which can be answered by the information from semi-structured data like tables and lists.}\label{fig:example}
\end{figure}

Question answering (QA) has become an important feature in most search engines as it delivers information to users in an effective and easy-to-understand manner. Answers to questions are often extracted from Web tables and lists.  For example, Figure~\ref{fig:example} shows the search result page (SERP) of questions $Q_1$ ``\emph{cities with the highest GDP in the world}'' and $Q_2$ ``\emph{the best skydiving locations in the world}'', where the answer to $Q_1$ is from a Web table, while that to $Q_2$ is from a Web list. 

Comparing to unstructured plain text, semi-structured Web data, such as Web tables and lists, are more effective to represent rich relational information. Relations among various elements in a Web table or a list may be useful in answering user questions.  According to the statistics from a global commercial search engine, there are hundreds of millions of semi-structured data pieces, including tables and lists on the Web, and the intents of $30\%$ of user queries can be answered by semi-structured data.

Previous attempts towards question answering (QA) using semi-structured data on the Web are mainly IR-based approaches~\cite{balakrishnan2015applying,chakrabarti2020open}.  Typically, those methods convert semi-structured data into documents by sequentially rearranging text cells to adapt to language models~\cite{chakrabarti2020open,wang2018neural,zhang2018ad}. 
Those studies do not make use of inherent structural relationships among components of Web tables or lists. For example, the rearrangement does not consider the vertical relations among cells locating in the same columns, such as the relation among ``New York'', ``Tokyo'' and ``Los Angeles'' in Figure~\ref{fig:example}(a).

Some recent studies leverage tabular structure implicitly. For example, \newcite{nishida2017understanding} cast tables as matrices of text and apply convolutional neural networks for table embedding. \newcite{zhang2018ad} cut tables into smaller fragments. However, the structural information in Web tables and lists is more complex than the simple adjacency relation of matrices. How to take the best advantage of both the text information and the structural relations in Web tables and lists in QA remains a challenge not thoroughly explored.

In this paper, we tackle the problem of Web QA over semi-structured data, and make the following contributions.
First, in Section~\ref{sec:problem}, we systematically categorize different components in semi-structured Web data, including captions, headers, subject columns, attribute columns, and cells, as well as their relations, including cell-cell relation, header-cell relation, subject-attribute relation, and caption-content relation. We propose \mname, a graph model to jointly represent both text and structural information in semi-structured data for Web QA in Section~\ref{sec:method}. Our \mname model explicitly represents different types of components as nodes in a graph and their relations as edges. Our model integrates heterogeneous information effectively, including text and structures, and reveals hidden semantic correlations across various components naturally.

Second, in Section~\ref{sec:method}, we apply two pre-training techniques for graph models. In particular, we design a novel node prediction objective (NPO) to leverage graph structure in node embedding. This pre-training task requires a model to predict the entire content for a masked node from the unmasked neighbor nodes, which guides the model to learn attention over the context.
The attention is used in the graph reasoning stage, where the representation of each node is updated by the aggregated information from the neighbor nodes. In this way, the inherent semantic correlations among neighbor nodes are propagated via the structural connections in the graph. Comparing to the previous methods, this graph pre-training and reasoning mechanism better exploits structural information in tables and lists.

Last, to compare our model with the state-of-the-art methods, we create new datasets which contain real-world queries collected from a global commercial search engine paired with table and list data mined from the Web. Each pair $\langle$query, table/list$\rangle$ is further labeled by crowdsourcing annotators with consensus on relevance. The experimental results on the test sets, reported in Section~\ref{sec:exp}, show that \mname outperforms the best state-of-the-art baselines up to 1.77 in F1 score on average.

\section{Problem Statement} \label{sec:problem}

Let $S$ be a semi-structured data example, either a Web table $T$ or a Web list $L$. There are different types of tables and lists, for example,  relational tables, entity tables, matrix tables, enumerate lists and group lists~\cite{lautert2013Web}. Our method is generally applicable to all those types, therefore, we do not distinguish them in this paper.

\begin{figure}[t] 
	\centering 
    \includegraphics[trim={1cm 3.3cm 1cm 7cm},clip,scale=0.45]{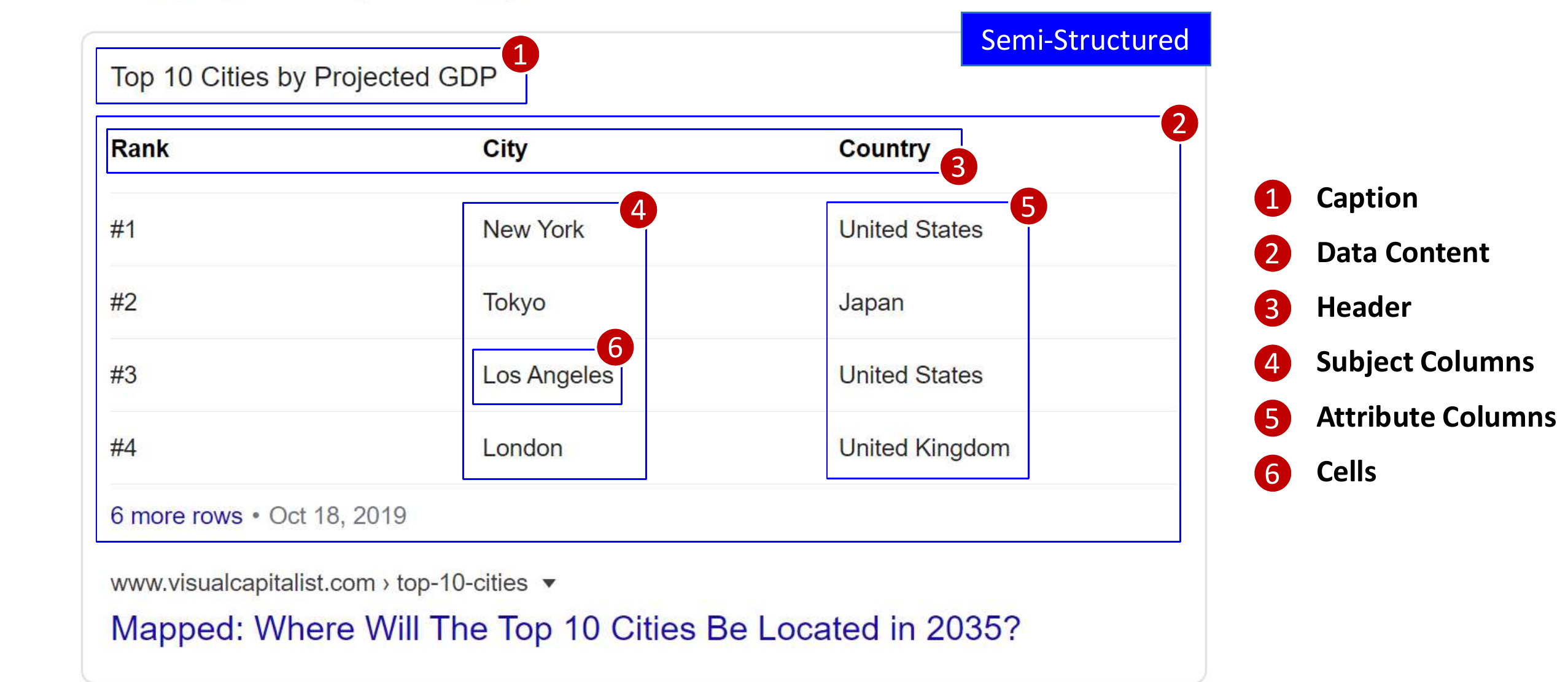}
    \caption{Components of Web Semi-Structured Data.}
	\label{fig:semi-component}
\end{figure}

\subsection{Components of Web Semi-structured Data}\label{sec:components}

Following~\newcite{crestan2011Web} and~\newcite{eberius2015building}, we divide a Web semi-structured data example into various components, as illustrated in Figure~\ref{fig:semi-component}. 

A \textbf{caption} $C$ is a direct description that is usually adjunct to the content body of the semi-structured data. For example, in Figure~\ref{fig:example} ``Top 10 Cities by Projected GDP'' and ``30 Jaw Dropping Places for Skydiving in the world!'' are captions for the table and the list, respectively.

\textbf{Data content} refers to the body of semi-structured data, which consists of multiple rows and columns. 
A special row, the \textbf{header}, often locates at the top of the table. The elements in a header often describe the classes that the content of the table belongs to. For example, in Figure~\ref{fig:semi-component}, the first row consisting of ``Rank'', ``City'', and ``Country'' is the header of the table. The elements in the remaining rows of the table are \textbf{cells}. Vertically, cells are grouped into columns, where we identify subject columns and attribute columns. \textbf{Subject columns} refer to one or more key subjects or entities of the table, while \textbf{attribute columns} list the attribute information of the corresponding subjects or entities. In Figure~\ref{fig:semi-component}, the column of ``City'' is a subject column, while the columns of `Rank'' and ``Country'' are both attribute columns. To recognize subject columns, we adopt a heuristic method~\cite{nishida2017understanding}, which calculates the distinct string ratio as seeds for subject classification. Our empirical study finds that this simple method achieves an accuracy over $95\%$. Besides, a schema classification method~\cite{eberius2015building} is applied to detect and transpose vertical Web tables into horizontal ones. 

A list can be regarded as a special type of table, which has only one single column, and has no header.

\subsection{Relations among Components in Tables and Lists}

Different components in tables and lists bear inherent semantic relations. Modeling those relations in a graph model fusions semantics among components and achieves a rich representation of semi-structured data. Particularly, we are interested in the following four types of relations.

\begin{description}
\item \textbf{Caption-Content Relation.} A caption is often a summary of the context and content in a table or a list. The words in a caption are often reliable evidences to determine the relevance between a query and a semi-structured data example.  

\item \textbf{Header-Cell Relation.} Since a header often outlines the classes that the cells belong to, a header-cell relation is usually a class-instance relation. For example, the cell, ``Los Angeles'' in Figure~\ref{fig:example} is an instance of the class ``City''.

\item \textbf{Subject-Attribute Relation}. More often than not tables store entity information. In such a table, each row, except for the header, corresponds to one entity, where the cells in the subject columns contain the entity names, and the remaining cells in the attribute columns consist of the attributes for that entities. For example, in Figure~\ref{fig:example}(a), the third row corresponds to a ``City'' entity ``Los Angeles'', and ``\#3'' and ``United States'' are the values of attributes ``Rank'' and ``Country'' of the entity, respectively. The subject-attribute relation is usually an entity-attribute relation.

\item \textbf{Cell-Cell Relation.} If we ignore the subject columns, the remaining cells within the same rows or columns are also semantically related. The cells in the same row often describe the various attributes of the same entity, while the cells in the same column are often instances of the same class.

\end{description}

As mentioned in Section~\ref{sec:components}, lists can be considered as a special type of tables. They only have Caption-Content relation and Cell-Cell relation. 

The problem of QA over semi-structured data is that, given a query $Q$ and a semi-structured data example $S$, return the {\bf QA match score} $d(Q,S)$, which predicts the likelihood that $S$ answers $Q$. 

\section{Method}\label{sec:method}

\begin{figure}[t]
\centering
\includegraphics[scale=0.22]{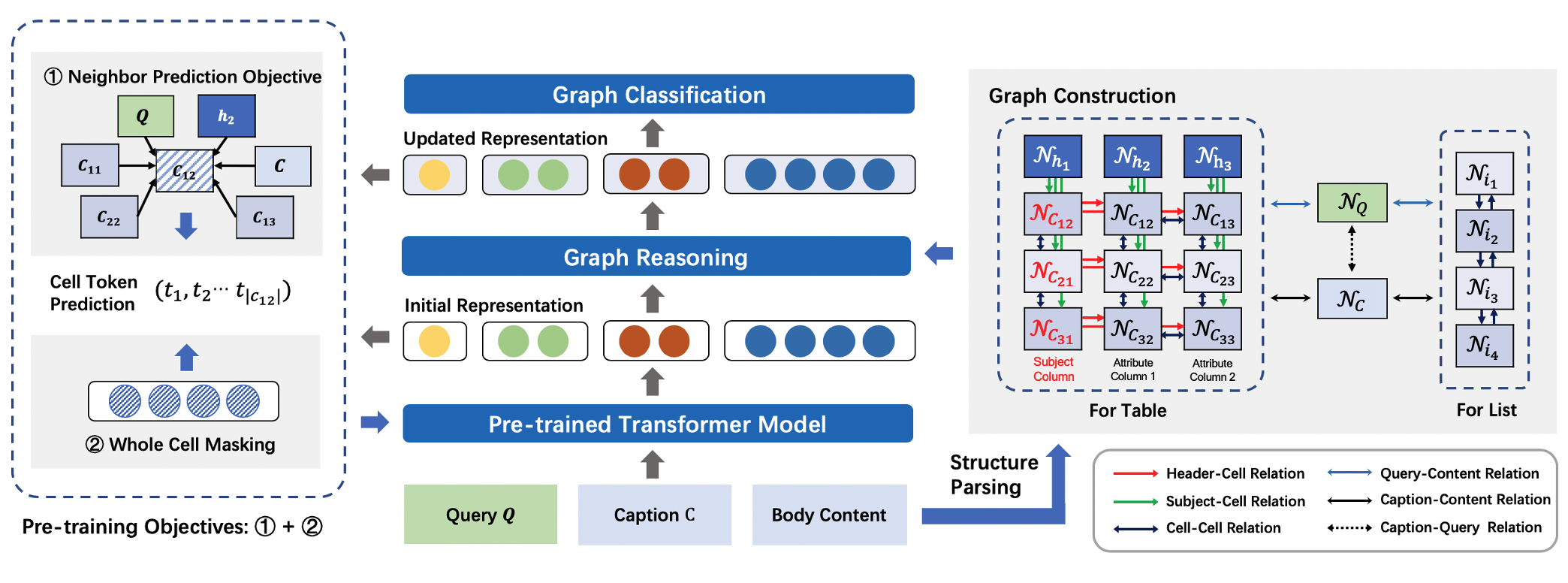}
\vskip -1em
\caption{\mname contextually encodes  query and semi-structured data via a pre-trained transformer model. It then builds a graph for the semi-structured data into a graph, and updates the node embeddings by the graph reasoning module. Finally the graph classification module aggregates the nodes information to predict the QA match score. In addition, \mname applies two graph pre-training objectives to encourage the model to learn attention over contextual nodes.
}\label{fig:architecture}
\end{figure}

In this section, we proposed \mname, which is a graph model of semi-structured data on the Web for QA.  Figure~\ref{fig:architecture} shows the overall structure of \mname.

\mname is composed of three components.  First, a \emph{pre-trained language model} generates token-level contextual embedding for the concatenation of an input query $Q$ and a semi-structured data example $S$. Second, a \emph{graph construction module} converts the initial plain text embedding into graphs. Last, a \emph{graph reasoning and classification module} predicts matching results. 

\subsection{Graph Construction}

Given a query $Q$ and a semi-structured data example $S$ of $M$ rows and $N$ columns, we construct a graph based on the components and their relations described in Section~\ref{sec:problem}. Figure~\ref{fig:architecture} illustrates the graph constitution:
\begin{mylist}
  \item \textit{Query Node} $\mathcal{N}_Q$; 
  \item \textit{Caption Node} $\mathcal{N}_C$;
  \item \textit{Header Nodes} $\{\mathcal{N}_{h_j}\}$, where $1\le j\le N$;
  \item \textit{Cell Nodes}  $\{\mathcal{N}_{c_{ij}}\}$, where $2\le i\le M, 1\le j\le N$.
\end{mylist}

The edges in the graph are created as follows. The first group of edges are formed based on the structural relations in the semi-structured data example. These structural relations carry the inherent semantic relations between the components, and define the context to better represent the elements in a semi-structured data example through the following four types of edges.
\begin{mylist}
  \item Caption-Content Relation: edges between caption nodes and cell nodes. 
  \item Header-Cell Relation: edges between header nodes and the cell nodes in the corresponding columns.
  \item Subject-Attribute Relation: edges between subject cell nodes and the attribute cell nodes in the corresponding rows.
   \item  Cell-Cell Relation: edges between neighbour cell nodes in the same row or in the same column.
\end{mylist}

The second group of edges connect the query and the semi-structured data example by connecting $\mathcal{N}_Q$ with all nodes in $S$. The weights of these edges will be derived in the graph reasoning stage to represent the bi-directional attention between the query words and data components, including the edges between query node and cell node, as well as the edges between query node and caption node.

The graph for a list is a simplified version of that for a table, where there are no Header-Cell edges or Subject-Cell edges.

\subsection{Graph Initialization}\label{sec:contextual_emb}

To obtain the initial representation of graph nodes, we first concatenate the query $Q$ and the text in the semi-structured data example $S$. The concatenated string consists of $G = (Q,C, \{h_j\}, \{c_{ij}\})$, where $C$ is the caption, $\{h_j\}$ are the tokens in the header, and $\{c_{ij}\}$ are the cells in $S$. We feed this concatenated string $G$ into a pre-trained BERT model~\cite{vaswani2017attention} and derive a contextual embedding for each token in $G$. We use $\mathcal{LM}(G)$ to denote this representation. In this paper, BERT\textsubscript{base}~\cite{devlin2018bert} is used for contextual embedding. 


We further derive the initial representation for each node in the graph. Since different nodes may contain various lengths of token spans, we adopt the method in~\cite{fang2019hierarchical}, which applies a BiLSTM~\cite{chen2017improving} on top of the transformer output and a multi-layer perceptron $MLP$ to convert various lengths of token spans into a fix-sized vector as the node representation. We write the BiLSTM model as a function $\mathcal{B}$. We denote by $\mathcal{B}(\mathcal{LM}(G))$ the model on top of the transformer output, and by $\mathcal{B}(\mathcal{LM}(G))[s;t]$ the sequence of hidden states in the model for span extremes in position $s$ and $t$.  We use the subscripts $start$ and $end$ to denote the start and end positions of the tokens of the corresponding components. The initial representation for the nodes are as follows, where normal fonts are used for the text of the corresponding nodes, and bold fonts for the embedding. 
\begin{align}\nonumber
    &\bm{Q} = MLP \Big (\mathcal{B}(\mathcal{LM}(\bm{G}))[Q_{start};Q_{end}] \Big) & 
    &\bm{C} = MLP\Big(\mathcal{B}(\mathcal{LM}(\bm{G}))[C_{start};C_{end}] \Big)\\
    &\bm{h_j} = MLP\Big(\mathcal{B}(\mathcal{LM}(\bm{G}))[h_{j;start};h_{j;end}] \Big) &
    &\bm{c_{i,j}} = MLP\Big(\mathcal{B}(\mathcal{LM}(\bm{G}))[c_{i,j;start};c_{i,j;end}] \Big) \nonumber
\end{align}

\subsection{Graph Reasoning and Prediction}

After generating the dense representation of graph nodes, \mname leverages a two-layer graph convolutional network (GCN)~\cite{kipf2016semi} to perform message passing over the graph. At each layer, the graph convolutional neural network aggregates the neighbors’ representations of one node and further transforms the aggregated representation with a linear projection. Let $\bm{L}^{(0)}=\{\bm{Q},\bm{C},\{\bm{h_j}\},\{\bm{c_{ij}}\}\} \in \mathbb{R}^{K \times d}$, where $K = 2+M\times N$ is the total number of nodes in the graph including the query node, caption nodes, header nodes and cell nodes, and $d$ is the output dimensionality of the MLP in Section~\ref{sec:contextual_emb}. The graph reasoning process is formalized as
\begin{align}\nonumber
    &\bm{L}^{(l)} = \sigma\Big(\tilde{\bm{D}}^{-\frac{1}{2}}\tilde{\bm{A}}\tilde{\bm{D}}^{-\frac{1}{2}}\bm{L}^{(l-1)}\bm{W}^{(l-1)}\Big),
\end{align}
where $\bm{L}^{(l)}$ denotes the $l$-th ($l=1,2$) layer of GCN, $\sigma$ is the non-linear activation function, which is ReLU in our case and $\bm{W}^{(l-1)}$ is the weight matrix of the $(l-1)$-th layer. $\bm{D} \in \mathbb{R}^{K \times K}$ denotes the  graph degree matrix, which records the amount of edges for every node and $\bm{A} \in \mathbb{R}^{ K \times K}$ denotes the graph adjacency matrix, which records the graph edge information. Symbol $\sim$ here indicates a renormalization trick of adding a self-connection to each node of the graph and building the corresponding degree and adjacency matrix. After two rounds of convolution, $\bm{L}^{(2)}$ denotes the node features updated.


Graph prediction is derived by a mean pooling operation on the nodes of the graph, followed by an MLP, that is,
$y = MLP(Pooling(\bm{L}^{(2)}))$, 
where $y$ is the predicted QA match score for the corresponding input query $Q$ and data example $S$.

\subsection{Pre-training Strategy}

Pre-training~\cite{erhan2010does} has become a new paradigm of natural language processing, and various pre-training techniques have been proposed~\cite{devlin2018bert,joshi2020spanbert}. However, most previous pre-training techniques were designed for plain text. Due to the structural characteristics of Web semi-structured data, they cannot be applied directly to such data. In this paper, we propose a novel pre-training method that allows the model to learn representations from semantics embedded in both text and structures of tables and lists. Following the successful pre-training experience of transformer-based models~\cite{devlin2018bert}, we used two pre-training objectives designed specifically for semi-structured data.

\textbf{Whole Cell Masking (WCM)}. We follow the masked language model proposed by BERT ~\cite{devlin2018bert}, but with different masking schema. Extended from whole word masking~\cite{joshi2020spanbert}, \emph{Whole Cell Masking} firstly masks every token of the word if any of its pieces is masked. Additionally, it masks the whole cell content if any token in table cells or headers is masked. We mask $15\%$ of all cells in total through replacing $80\%$ of the masked cell tokens by a special mask token [MASK], $10\%$ by random tokens and $10\%$ with the original tokens. Given input $G = (Q,C, \{h_j\}, \{c_{ij}\})$, let $T = (t_1,\ldots, t_{|X|})$ be the sequence of tokens for $G$, where $t_m \in T$ is the $m$-th token, which is masked, that is, $\bm{t}_m = MLP(\bm{e}_m)$,
where $\bm{e}_m \in \mathbb{R}^{d}$ denotes the token-level embedding of input $t_m$, which is generated by the contextual language model. After an MLP with one hidden layer, $\bm{e}_m$ is decoded as a token prediction score $\bm{t}_m \in \mathbb{R}^{V}$, where $V$ denotes the vocabulary size.

\textbf{Neighbor Prediction Objective (NPO)}. To incorporate the structural information of semi-structured data in the pre-training stage, we propose a novel \emph{neighbor prediction objective} task for graph pre-training. The task is to predict each token inside a masked node using the representations of the neighbor nodes. In order to make the pre-training consistent with fine-tuning stage, we apply the same contextual embedding module and graph reasoning module as used in the fine-tuning process to generate the reasoned node representation.

Formally, denote by $\bm{L}^{(2)}_{n}$ the node representation of the $n$-th node $\mathcal{N}_n$ after contextual embedding and graph reasoning, and by function $neighbour(\cdot)$ the node representations of its neighbor nodes in the constructed graph. We use fixed sinusoidal embedding~\cite{liu2019roberta} as positional embedding to predict the tokens from $\bm{L}^{(2)}_{n}$.
\begin{align}
    &\bm{r}_{n_k} = [Pooling(neighbour(\bm{L}^{(2)}_{n}));\bm{p}_k]\\
    &\bm{t}_{n_k} = MLP(\bm{r}_{n_k})
\end{align}
where function $Pooling(\cdot)$ converts the neighbor node representations of $\bm{L}^{(2)}_{n}$ into a $d$-dimensional vector with mean pooling. We concatenate the representations of the neighbors and the $k$-th positional embedding $\bm{p}_k \in \mathbb{R}^{d}$ to get the representation for the $k$-th token $\bm{r}_{n_k}$. After an MLP with two hidden layers for decoding, we obtain the prediction result of the $k$-th token $\bm{t}_{n_k} \in \mathbb{R}^{V}$.

\mname sums up the loss from both the whole cell masking objective and the neighbor prediction objective as the total loss function. For the $m$-th token in the input token sequence, we can find the corresponding position $k$ in the $n$-th node. The total pre-training loss is
\begin{align}
    &\mathcal{L}(t_m) = \mathcal{L}_{WCM}(t_m) + \mathcal{L}_{NPO}(t_m) = -log P(t_m|\bm{t}_m) - log P(t_m|\bm{t}_{n_k})
\end{align}

Notably, NPO directly uses the masked input from WCM for graph construction and prediction.

\section{Experiments}\label{sec:exp}

We evaluate the \mname model and other baselines on three datasets, including one table QA dataset, one list QA dataset and one small dataset of complex query-table pairs. In addition, we leverage two other large-scale datasets for pre-training. We describe the three datasets as follows.
\begin{itemize}

\item \textbf{Table Query Matching dataset (Table-QM)} is
an English table QA task dataset from one commercial Q\&A system, which has about 34k labeled cases. Each case consists of three parts, a question, a table, and a binary label (0 or 1) by crowdsourcing judges indicating whether the question can be answered by the table. The dataset is collected as follows. First, for each question, the top 10 relevant documents returned by the search engine are selected to form pairs $\langle$Question, Url$\rangle$. Then, tables are extracted from those documents to form triples $\langle$Question, Url, Table$\rangle$. Those $\langle$Query, Table$\rangle$ pairs are sampled and sent to crowdsourcing judges. Specifically, each pair $\langle$Query, Table$\rangle$ is required to be labeled by three judges. Those cases with 2/3 or higher positive labels receive positive final labels, otherwise negative. 

\item \textbf{List Query Matching dataset (List-QM)} is an English list QA task dataset from one commercial Q\&A system, which has about 62k labeled cases. The data collection process is similar to Table-QM. For query selection, we include unordered lists, ordered lists and description lists~\cite{world1999html} to increase diversity.

  \item  \textbf{Deep Tables Query Matching dataset (DTable-QM)} is a subset of the Table-DM dataset, including those tables with relatively complicated structures and larger amounts of information. Specifically, we selected the tables with more than 50 cells and 10 columns, which may bring challenges for semantic table modeling. This results in 2,457 instances in the dataset. \end{itemize}

For pre-training of \mname, we leverage the following semi-structured datasets.  
\begin{itemize}
\item \textbf{Large-scale Table Pre-training dataset (Table-Pretrain)} is a large-scale unlabelled Web table dataset, which is extracted from 10 million Web pages sampled from the index of one commercial search engine. The sampling is conducted from the set of pages seen by United State users in the period from Feb 2019 to Dec 2019. After filtering out incomplete or empty Web tables, we extract 7M Web tables.

\item \textbf{Large-scale List Pre-training dataset (List-Pretrain)} is a large-scale unlabelled Web list dataset. Similar to the process of deriving Table-Pretrain, 100M Web lists are extracted.
\end{itemize}

\begin{table}[t]
\begin{center}
\small
\begin{tabular}{l|ccccc}

\toprule
\textbf{Statistic}& Table-QM & List-QM & DTable-QM & Table-Pretrain& List-Pretrain  \\
\midrule
Dataset size& 34,276 & 62,581 & 2,457& 6,994,996 & 100,954,174 \\
Avg. query length  & 28.0 & 30.5 & 28.3 & 27.9 & 30.5  \\
Avg. title length  & 134.1 & 34.9 & 124.3 & 103.4 & 30.5  \\
Avg. \# of rows  & 7.6 & 5.8 & 10.3  & 4.7 & 11.4  \\
Avg. \# of columns  & 2.8 & 1 & 5.6 & 2.4 & 1  \\
Avg. \# of cells & 21.3 & 5.8 & 57.7 & 11.3 & 11.3  \\
Positive/Negative & 1:2.1 & 1:1.3 & 1:2.1 & - & - \\
\bottomrule
\end{tabular}
\end{center}
\caption{\label{data-table} Dataset Statistics. }
\end{table}

We compare \mname with several strong baselines. \textbf{Single-field document retrieval (SDR)}~\cite{cafarella2009data,cafarella2008Webtables} and \textbf{Multi-field document retrieval (MDR)}~\cite{pimplikar2012answering} are two representative methods that treat a semi-structured data example as a single document or multi-fielded document. They apply an IR approach for QA~\cite{zhang2018ad}.  \textbf{Semantic table retrieval (STR)}~\cite{zhang2018ad} introduces a semantic representation for Web tables. The representation includes sets of extracted concepts and entities. \textbf{BERT}~\cite{devlin2018bert} is a powerful Transformer-based model, which has demonstrated impressive performance in the semantic matching task. We apply this model to a concatenation of the query and the sequential tokens in a semi-structured data example, and then use a multi-layer perception for classification. All the previous methods do not consider the structure information in tables or lists. In this paper, we applied Bert-base as our backbone model and baseline. The last baseline is \textbf{TAPAS}~\cite{herzig2020tapas}, a recent state-of-the-art approach of QA over tables. This method encodes rows and columns to embed structural information of tables. 

To measure the accuracy of matching, we use the average F1 as our metric. Precision, Recall, and F1 score are computed on the number of true positives (TP), false positives (FP), and false negatives (FN). F1 score is the harmonic mean of precision and recall. Since the matching prediction task is casted as a binary classification task, we consider F1 score as the metric and calculate average F1 based on that.

All methods are implemented in PyTorch~\cite{paszke2017automatic} and trained on an Ubuntu 16.04 with 64GB memory and eight GTX 1080 Ti GPU. For all data-sets, we randomly select 80\% of the records as training set, 10\% as validation set and the remaining 10\% as test set. We train the model using training data, and fix model parameters based on the best model performance on validation set. We then test the model on test set. We perform three random runs and report both mean and standard deviation for testing performance. 

We use stochastic gradient descent (SGD) with a learning rate of 2e-5. We use mini-batches of size 64, with batch size 8 for each of 8 GPUs, we use with 1 hidden-layer of 768 hidden units. We use dropout with a rate of 0.5, which is applied to all feedforward neural networks. For the pre-trainng process, We use a batch size of 64 and fine-tune for 4 epochs over the large-scale data-set for two unsupervised task. For each task, we selected the fine-tuning learning rate of 2e-5. For the graph convolutional network, we applied a Bi-LSTM with hidden-layer with 768 hidden units on the top of transformer output. The GCN contains two convolutional layers with the hidden size of 1,536. After node-level convolution, we adapted mean-pooling for graph representation. As to the positional embedding, we created a fixed sinusoidal embedding with 768 hidden units.

For all baseline models, we use pre-trained corresponding transformer models as word embedding and using the output of token [CLS] as sentence embedding. Out-of-vocabulary (OOV) words are hashed to one of 100 random embedding each initialized to mean 0 and standard deviation 1. All other hidden layer weights were initialized from random Gaussian distribution with mean 0 and standard deviation 0.01. Each hyperparameter setting was run on a same machine as the \mname, using Adagrad for optimization with initial accumulator value of 0.1.

\subsection{Overall Performance}

We compare \mname against the state-of-the-art baselines on the Table-QM, List-QM and DTable-QM datasets. The results are reported in Table~\ref{performance-table}. As \mname is complementary to language models, we use \mname(BERT) to denote the language models used by \mname as the backbone model.

\begin{table}[h]
\begin{center}
\begin{tabular}{l|cccc}
\toprule 
\bf Method & \bf Table-QM & \bf List-QM & \bf DTable-QM  & \bf Average \\ \hline
SDR~\cite{cafarella2009data} & 64.39  & 62.94  & 57.95 & 61.76\\
MDR~\cite{pimplikar2012answering} & 65.82 & 62.79 & 59.21 & 62.61 \\
STR~\cite{zhang2018ad} & 72.95 & 69.93 & 65.30 & 69.39 \\
BERT~\cite{devlin2018bert} & 76.22 & 72.15 & 68.54 & 72.30 \\
TAPAS~\cite{herzig2020tapas} & 77.92 & 73.69 & 70.49 & 74.03 \\
\midrule
\mname & \bf 79.04 & \bf 77.61 & \bf 71.19 & \bf 75.95 \\
\bottomrule
\end{tabular}
\end{center}
\caption{\label{performance-table}Performance comparison on the three datasets. The \% signs are omitted. The best results are highlighted in bold.}
\end{table}

\mname consistently achieves the best performance against all baselines. \mname outperforms the baselines BERT by up to $5.44\%$ (List-QM). Our model captures both the text-level and structure-level information via explicitly modeling the inherit building components and their semantic correlation from Web semi-structured data. Comparing to the best IR-based method STR, our model is up to 7.68\% better on the List-QM dataset. It demonstrates that the heterogeneous graph model in \mname uses structural features more effectively than those IR-based methods, which focus on slicing Web semi-structured data into different documents but ignore the potential correlations among them. Besides, \mname outperforms TAPAS, the newest baseline for QA on tables, by up to 3.90\% in the List-QM dataset. It illustrates that the graph-based pre-training objectives strengthen the representation capability of models for semi-structured data, which will be further discussed in Section~\ref{sec:pretrain-analysis}. 

Notably, all the baselines display servere performance drops in the DTable-QM dataset, while \mname still holds the best performance (71.19\%). The explicit graph modeling guides the model to learn attention over noisy contexts, which benefit semantic reasoning on complicated tables.

\subsection{Ablation Studies}

We conduct ablation studies on \mname to empirically examine the contribution of every components, particularly, the semantic relations we proposed, which includes the following steps.

\noindent\textbf{Semantic Relation Ablation} To further study the contribution of the semantic relations defined in Section 2.2, we removed the edges representing \textit{Caption-Content Relation}, \textit{Header-Cell Relation}, \textit{Subject-Attribute Relation} and  \textit{Cell-Cell Relation} from graphs respectively and keep the other components untouched.

\noindent\textbf{LSTM Ablation} We replace Bi-LSTM, which generates node representation from the output of language model, by average pooling to obtain the fix-sized initial embedding as the inputs of graph neural network. 

\noindent\textbf{GCN Ablation} We remove GCN, which aggregates and updates node-level representation and outputs the final prediction. We also remove the LSTM part as there is no need to generate node inputs. Instead, we use a MLP for classification, which makes the model same as one of our baselines BERT.

\begin{table}[h]
\begin{center}
\small
\begin{tabular}{l|cccc}
\toprule 

\bf Method & \bf Table-QM & \bf List-QM & \bf DTable-QM &\multirow{2}{*}{ Average}\\ \cline{1-4}

\mname & 79.04 & 77.61 & 71.19 \\
\hline

w/o Caption-Content Rel. &  78.55 (0.49 $\downarrow$)  & 76.04 (1.57 $\downarrow$) & 70.90 (0.29 $\downarrow$)& 0.78$\downarrow$ \\
w/o Header-Cell Rel. & 77.37 (1.67 $\downarrow$)  & -  & 70.21 (0.98 $\downarrow$) & 1.33$\downarrow$ \\
w/o Subject-Attribute Rel. &  78.52 (0.52 $\downarrow$)  & -  & 70.58 (0.61 $\downarrow$)& 0.57$\downarrow$ \\
w/o Cell-Cell Rel. & 77.92 (1.12 $\downarrow$)  & 74.60 (3.01 $\downarrow$) & 70.10 (1.09 $\downarrow$)& 1.74$\downarrow$ \\
w/o LSTM & 78.34 (0.79 $\downarrow$)  & 76.16 (1.55 $\downarrow$) & 69.90 (1.29 $\downarrow$)& 1.21$\downarrow$ \\
w/o GCN & \bf 76.30 (2.74 $\downarrow$) & \bf 72.49 (5.12 $\downarrow$) & \bf 68.97 (2.22 $\downarrow$)& \bf 3.36$\downarrow$\\

\bottomrule
\end{tabular}
\end{center}
\caption{\label{ablation-table}Ablation studies on the main components, where the last column shows the average of performance reduction. The numbers are in percentage and the signs $\%$ are omitted.}
\vskip -0.5em
\end{table}

The results show that ablation causes performance degrade to different extents. We can observe that removing GCN, which conducts explicit graph reasoning, causes serious performance dropping 3.36\% on average. It again confirms the effectiveness of explicitly modeling the inherit building components and their semantic correlations from Web semi-structured data, especially for lists (a decrease of 5.12\%). The semantic relation ablations elaborate on the contribution of each relations in semantics fusion among components:  Among them, the Cell-Cell relations is proven to contribute most in semantic modeling for its largest performance reduction(1.74\% in average). The \mname w/o Header-Cell/Subject-Attribute Relation dropped 1.33\% and 0.57\%  in average, indicating the GCN successfully utilized these relations in table modeling. Additionally, the replacement of the Bi-LSTM component reduces the overall performance the least (0.79\% on average), but is still better than simply using pooling method for token aggregation.

\subsection{Pre-training Strategy Analysis}\label{sec:pretrain-analysis}

To evaluate the proposed pre-training techniques, we train the original \mname model with different objectives. Specifically, we apply only one pre-training objective each time and evaluate their performance on the three 
datasets. The evaluation results are shown in Table~\ref{pretraining-table}.

\begin{table}[h]
\begin{center}
\small
\begin{tabular}{l|ccc}
\toprule 
\bf Pre-training Strategy & \bf Table-QM & \bf List-QM & \bf DTable-QM  \\ \hline
None & 77.41 & 76.79 & 69.22 \\
\hline
WCM & 77.66 (0.25 $\uparrow$) & 77.29 (0.50 $\uparrow$) & 70.98 (1.76 $\uparrow$) \\
NPO & 78.54 (1.13 $\uparrow$) & 77.42 (0.63 $\uparrow$) & 71.06 (1.84 $\uparrow$) \\
WCM+NPO & \bf 79.04 (1.63 $\uparrow$)& \bf 77.61 (0.82 $\uparrow$)& \bf 71.19 (1.97 $\uparrow$)\\
\bottomrule
\end{tabular}
\end{center}
\caption{\label{pretraining-table}Performance comparison of pre-training objectives. The metrics are in percentage with signs \% omitted.}
\end{table}

Each objective contributes to the performance improvement. When we solely use WCM as the pre-training objective, the performance is increased up to 1.97\% on all three datasets. WCM successfully guides the model to learn reasonable token-level embedding. 
Via pre-training with NPO only, the performance is increased up to 1.84\%. NPO allows the inherent semantic correlations among neighbor nodes to be propagated via the structural graph connections. The combination of WCM and NPO achieves the largest performance increase (1.97\%), showing that this pre-training strategy exploits the structural information in tables and lists the best.

\section{Related Work}

Early studies on Query-Table Matching adopt IR approaches. For example, \newcite{chakrabarti2020open} and~\newcite{pimplikar2012answering} convert Web tables into multi-field documents and apply document retrieval pipelines proposed in~\cite{jurafsky2006speech,pacsca2003open}. \newcite{zhang2018ad} propose to create semantic features at text level, concept level and entity level. These methods mainly consider the textual information in tables, but largely ignore the inherent structural information in tables. 


In recent years, learning representations for semi-structured data has received increasing interest. \newcite{nishida2017understanding} propose to apply convolutional models to Web table. The rationale is to consider a Web table as a matrix of text, analogous to an image of pixels. However, their model does not show strong performance, partly because the semantic relationship among neighbor cells in tables may be far more complex than the simple adjacency relation among neighbor pixels in an image.  \newcite{herzig2020tapas} propose TAPAS, a weakly supervised table parsing method. TAPAS models the structure information of tables by explicitly encoding rows and columns. Similarly, \newcite{yin2020tabert} propose TABERT, which focuses on pre-training methods for the table QA task. The authors design a pipeline for learning row-level and column-level representations. \newcite{muller2019answering} also builds a graph representation on tables cells, focusing on optimizing cell answer selection. However, These works only model the row/column relations among table cells, without considering other relations including caption-content relation, header-cell relation and subject-attribute relation. In this work, we give a thorough categorization of the relations among all components in semi-structured data, and propose a graph model to incorporate all these relations. 

Our work is also generally related to the broad areas of graph neural networks and pre-training techniques, we refer interested readers to \cite{wu2020comprehensive} and \cite{qiu2020pre} for comprehensive surveys. 

\section{Conclusion}
Semi-structured data on the Web, including tables and lists, present a rich source for Web QA. Most of the previous methods do not take full advantage of structural information in semi-structured data.  In this paper, we propose a novel approach to model both textual and structural information in semi-structured data. Extensive experimental results verify the effectiveness of our approach.

\bibliographystyle{acl}
\bibliography{coling2020}

\end{document}